\documentclass[journal]{IEEEtran}

\usepackage{graphicx}
\usepackage[pagebackref=true,breaklinks=true,letterpaper=true,colorlinks,bookmarks=false]{hyperref}
\usepackage{amssymb}
\usepackage{multirow}
\usepackage{cite}
\usepackage{xspace}
\usepackage{subfigure}

\newcommand*{\etal}{\textit{et al.}\@\xspace}
\newcommand*{\ie}{\textit{i.e.}\@\xspace}
\newcommand*{\eg}{\textit{e.g.}\@\xspace}

\begin{document}

\title{Content-aware Directed Propagation Network with Pixel Adaptive Kernel Attention}

\author{Min-Cheol~Sagong, Yoon-Jae~Yeo, Seung-Won Jung,~\IEEEmembership{Senior Member, IEEE}, and Sung-Jea Ko,~\IEEEmembership{Fellow, IEEE}
\thanks{\textit{Corresponding author: Seung-Won Jung.}}
\thanks{M.-C. Sagong, Y.-J. Yeo, S.-W. Jung, and S.-J. Ko are with School of Electrical Engineering Department, Korea University, Anam-dong, Sungbuk-gu, Seoul, 136-713, Rep. of Korea (e-mail: mcsagong@dali.korea.ac.kr, yjyeo@dali.korea.ac.kr, swjung83@korea.ac.kr, sjko@korea.ac.kr).}}% <-this % stops a space

%\thanks{Manuscript received April 19, 2005; revised August 26, 2015.}}

% The paper headers
\markboth{}{Shell \MakeLowercase{\textit{Sagong et al.}}}

\maketitle

\begin{abstract}
   Convolutional neural networks (CNNs) have been not only widespread but also achieved noticeable results on numerous applications including image classification, restoration, and generation. Although the weight-sharing property of convolutions makes them widely adopted in various tasks, its content-agnostic characteristic can also be considered a major drawback. To solve this problem, in this paper, we propose a novel operation, called pixel adaptive kernel attention (PAKA). PAKA provides directivity to the filter weights by multiplying spatially varying attention from learnable features. The proposed method infers pixel-adaptive attention maps along the channel and spatial directions separately to address the decomposed model with fewer parameters. Our method is trainable in an end-to-end manner and applicable to any CNN-based models. In addition, we propose an improved information aggregation module with PAKA, called the hierarchical PAKA module~(HPM). We demonstrate the superiority of our HPM by presenting state-of-the-art performance on semantic segmentation compared to the conventional information aggregation modules. We validate the proposed method through additional ablation studies and visualizing the effect of PAKA providing directivity to the weights of convolutions. We also show the generalizability of the proposed method by applying it to multi-modal tasks especially color-guided depth map super-resolution. 
\end{abstract}

\begin{IEEEkeywords}
Deep learning, Content-adaptive convolution, Semantic segmentation, Color-guided depth map super-resolution
\end{IEEEkeywords}

\section{Introduction}
\IEEEPARstart{D}{eep} learning based on convolutional neural networks (CNNs) has brought remarkable improvement to image processing and computer vision tasks including object detection~\cite{fastrcnn, deformable, cornernet, Cao_2020_CVPR, Tan_2020_CVPR, ji2020casnet}, classification~\cite{vgg, resnet, inception, xie2020self, Zhang_2020_CVPR}, image restoration~\cite{pepsi, srcnn, compression, Zamir_2020_CVPR, Suin_2020_CVPR}, and generation~\cite{gan, style, Zhu_2020_CVPR, Lee_2020_CVPR, pepsi++}. Convolution is a basic element of CNNs, and has been considered as one of the most effective methods to extract and propagate features from images. Due to its weight sharing property, CNNs require less parameters than fully-connected layer and can be efficiently optimized by GPU implementation. However, the learned filters stay fixed after training in traditional convolution, making operation content-agnostic.

Toward content-adaptive convolution, a learnable filter which is generated dynamically conditioned on input features was proposed and showed promising effectiveness through the quantitative and qualitative performance evaluation~\cite{dfn, pac, deformable}. In particular, with the marginal increase in the number of network parameters, its adaptive property improves flexibility of the model. Moreover, Su~\etal~\cite{pac} introduced a pixel-adaptive convolution~(PAC) which multiplies a spatially-varying kernel with the shared filter weights. It can be ideally used for a wide range of tasks but its effectiveness was shown for only a few image filtering tasks. To further improve flexibility of convolution, a deformable convolution~\cite{deformable}, in which spatial sampling locations are augmented with additional learnable offsets, was proposed. This helps CNNs enhance the transformation modeling capability without additional supervision. To focus more precisely on pertinent image regions, the deformable convolution was further reformulated by modulating the input feature amplitudes according to the spatial locations and bins~\cite{deformable2}. This modulation can prevent features to be influenced by irrelevant content outside the region of interest. 

In this paper, we propose a novel convolutional operation, called pixel adaptive kernel attention~(PAKA), which drives the standard convolution to handle a content-adaptive receptive field. Specifically, PAKA modifies the weights of convolution with directional and channel modulations. The directional modulation emphasizes or suppresses features from different kernel directions while channel modulation aggregates the inter-channel relationship. With PAKA, the convolution predicts directions that provide pertinent information in every pixel. We evaluate the efficacy of PAKA with various experiments on multiple tasks. For the semantic segmentation task, we propose a hierarchical module with PAKA which can utilize diverse effective patch sizes. Through the proposed module, the network learns to attend to content-adaptive directions to inherit more information. To visually demonstrate this behavior, we define the modulated receptive field, called propagational field, which is emphasized or suppressed receptive field by the directional modulation. We validate our module by comparing ours with the state-of-the-art information aggregation modules under the same conditions. To demonstrate the generalizability of our method, we also apply PAKA to the joint up-sampling layer for the color-guided depth map super-resolution task. The experimental results indicate that the proposed method not only exhibits superiority on various CNN-based tasks but also has a noticeable potential.

In summary, in this paper we present:
\begin{itemize}
\item A novel convolutional operation that provides directivity to the standard convolution to address its limitation which is content-agnostic.
\item A novel information aggregation module and extensive experiments to validate the effectiveness of the proposed method on semantic segmentation.
\item Application of the proposed method to the joint up-sampling layer and extensive experiments to validate its effectiveness on color-guided depth map super-resolution.
\end{itemize}

\section{Related Works}
\label{sec:related}

\textbf{Content-adaptive filters} Recently, several works have explored the idea of utilizing effective content-adaptive filtering techniques such as bilateral filtering~\cite{bilateral, gaussian} and guided filtering~\cite{guided} as the layers of the CNNs. In the early stages in this direction, some approaches~\cite{structure, crf} made these filters differentiable to back-propagate the gradients for learning network parameters. Moreover, the learnable layers that play the role of the bilateral filter were applied to superpixels~\cite{superpixel}. On the other hand, by reformulating the guided filter as a fully differentiable block, Wu~\etal~\cite{end} proposed guided filtering layers that can be jointly optimized through end-to-end training. While the aforementioned methods cannot fully replace the standard convolution by their proposed layers, Jampani~\etal~\cite{sparse} introduced a sparse high-dimensional convolution that modifies the standard convolution to be content-adaptive. Similarly, Su~\etal~\cite{pac} presented a generalized convolution, called PAC, which can learn adaptive filters and has less computational overhead compared to~\cite{sparse}. In addition, introduced by Jia~\etal~\cite{dfn}, the dynamic filter network (DFN) directly predicts filter weights using a separate network branch; thus, it can generate adaptive filters corresponding to each input feature.

\textbf{Attention mechanisms} In learning a series of pattern recognition tasks, depending on the characteristic of each task, the network should reflect that given feature maps have different importance along spatial and/or channel direction. To this end, there have been several attempts to adopt the attention modules, which compute the responses for the local part while attending to the global context. Wang~\etal~\cite{residualattention} introduced a residual attention module which directly generates 3D attention map to refine the intermediate feature map. By using this module, the network performs robustly against noisy labels. Meanwhile, Hu~\etal~\cite{squeeze} proposed a squeeze-and-excitation module that computes channel attention with global average pooling. Even though the architecture effectively exploits the inter-channel relationships, they did not consider the spatial attention which plays an important role in inferring accurate attention for 3D feature maps. On the other hand, some approaches~\cite{bam, cbam} separately learn both channel attention and spatial attention. By applying the decomposed attention generation, the network showed superior performance than~\cite{squeeze} as well as much less parameter overhead than~\cite{residualattention}.

\textbf{Information aggregation modules}
In recent years, many researches have explored information aggregation for scene understanding. Zhao~\etal~\cite{pspnet} proposed PSPNet which adopts the pyramid spatial pooling~(PSP) module~\cite{psp} to reduce the feature maps into different scales. PSPNet utilizes various local context including the global information. On the other hand, Deeplab methods~\cite{deeplab1, deeplab2, deeplab3} introduced atrous spatial pyramid pooling~(ASPP) which applies sampling with different rates for information aggregation. Fu~\etal~\cite{dual} and Yuan~\etal~\cite{ocr} adopted the self-attention mechanism~\cite{self} to aggregate long-range spatial information, while Zhang~\etal~\cite{cem} utilized context encoding module~(CEM) which contains the global pooling to capture the global context and highlights the class-dependent feature maps.

\begin{figure*}[t]
\centering
\includegraphics[width=0.9\textwidth]{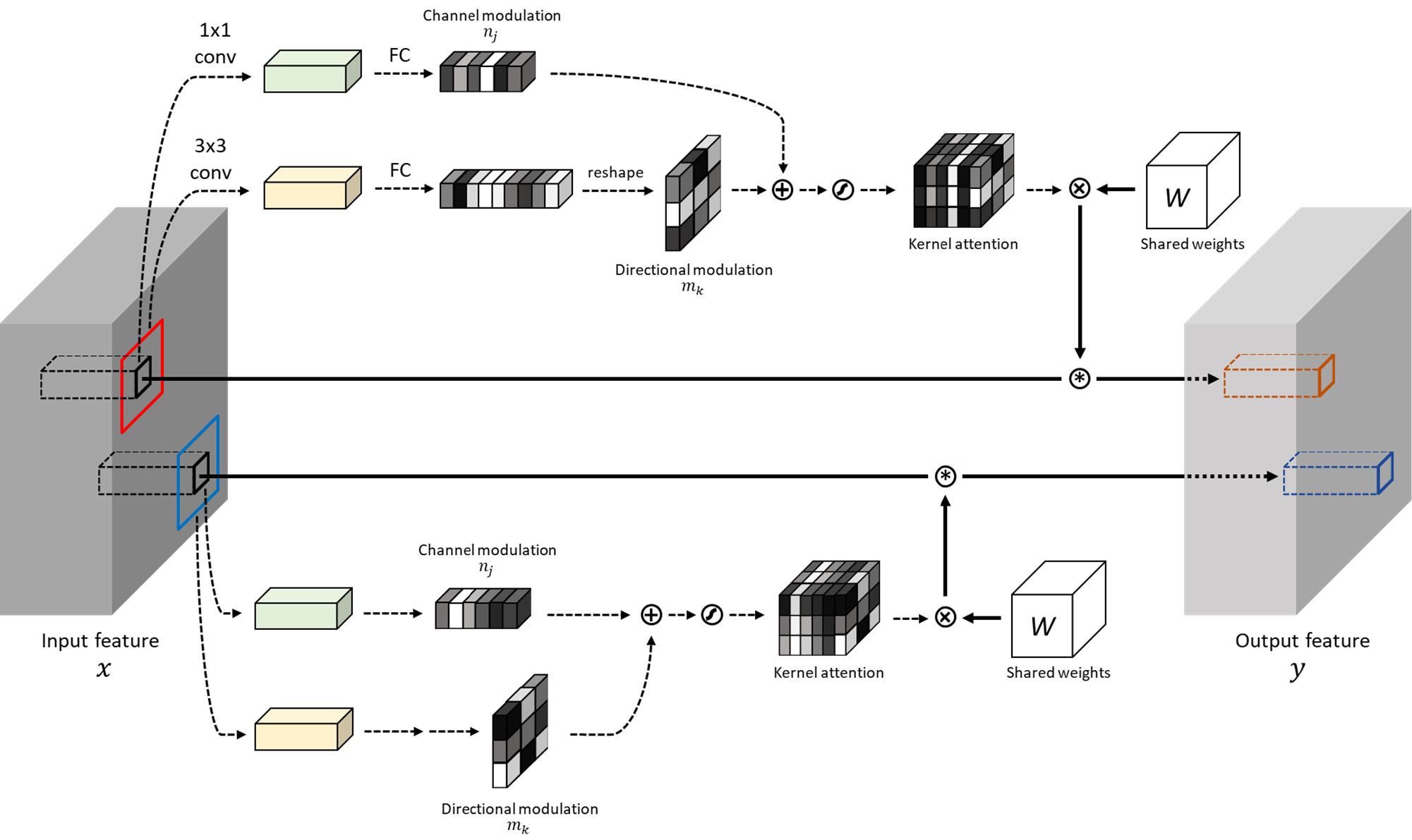}
\caption{Details of PAKA. Given the intermediate feature map $x$, PAKA computes the channel modulation~$n_j$ and directional modulation~$m_k$ through the two separate branches in every pixel. Both branches contain a couple of convolutional layers, the batch normalization, and ReLU. The two modulations are combined with the tensor broadcast and generate the kernel attention by the activation function. Since the kernel attention is different in every pixel, the proposed convolutional layer can learn the content-adaptive directivity.}
\label{fig:paka}
\end{figure*}

\section{Proposed Method}

Fig.~\ref{fig:paka} illustrates the modified convolution in which PAKA is applied. The proposed method produces kernel attention from input feature through two separate branches, \ie, channel modulation branch and directional modulation branch. In this section, we describe details of PAKA.

\subsection{Pixel Adaptive Kernel Attention}

Formally, convolution for each location $p$ on the output feature map $y \in \mathbb{R}^{H \times W}$ and input feature maps $x~\in~\mathbb{R}^{N \times H \times W}$ can be expressed as follows:

\begin{eqnarray}
\label{eqn:conv}
y(p) = \sum_{j=1}^{N}\sum_{k=1}^{K}x(p+p_k, j) \cdot w(k, j),
\end{eqnarray}
where $w(k, j)$ denotes the weight for the \textit{k}-th location and the \textit{j}-th channel, and $p_k$ is the pre-specified offset for the \textit{k}-th location. \textit{N} and \textit{K} are the numbers of channels and sampling locations, respectively. For instance, $K=9$ and $p_k~\in~\{(-1, -1), (-1, 0), \dots, (1, 1)\}$ correspond to $3\times3$ convolution with dilation 1. (\ref{eqn:conv}) indicates that the weights only depend on pixel and channel positions; thus, the standard convolution is content-agnostic. To cope with this problem, we designed PAKA which provides directivity to weights along pixel contents. With PAKA, (\ref{eqn:conv}) becomes

\begin{eqnarray}
y(p) = \sum_{j=1}^{N}\sum_{k=1}^{K}x(p+p_k, j) \cdot w(k, j) \cdot A_{k,j}(p),
\end{eqnarray}
where $A_{k,j}$ is learnable kernel attention for the \textit{k}-th location and the \textit{j}-th channel. The attention lies in the bounded range by applying the activation function to the combination of directional and channel modulations, which are denoted as $m_k~\in~\mathbb{R}^{K \times H \times W}$ and $n_j~\in~\mathbb{R}^{N \times H \times W}$, respectively. Each modulation is obtained by individual branches that apply multiple learnable layers to the same input feature maps~\textit{x}. By combining two decomposed modulations, PAKA can conduct attention mechanism in an efficient and effective manner. In particular, we apply element-wise summation with the tensor broadcast to combine them for efficient gradient flow~\cite{resnet}. As a result, the kernel attention can be expressed as

\begin{eqnarray}
A_{k, j} = 1 + \tanh(m_k + n_j).
\end{eqnarray}

\textbf{Channel modulation branch}
Each pixel in feature maps contains information from different contents (\eg, pixels from the object and background). As each channel tends to respond to a specific feature, we drive PAKA to exploit inter-channel relationship from every single pixel by channel modulation. We use a multi-layer perceptron including two $1\times1$ convolutional layers followed by a batch normalization layer to estimate channel modulation.

\textbf{Directional modulation branch}
Pixels corresponding to the same object can be surrounded by different objects (\eg, pixels in the center and around boundary of objects). However, since the standard convolution applies shared weights across every pixel, a trained network propagates information from the same directions for different pixels in an image. To solve this problem, the directional modulation $m_k$ is acquired to emphasize or suppress information from different directions. Specifically, it infers which kernel direction among the pre-specified offsets should be focused on every pixel. By adopting multiple convolutional layers with PAKA, each pixel can take information propagated from different fields. We design the directional modulation branch using a $3\times3$ convolutional layer followed by a $1\times1$ convolutional layer. In particular, the $3\times3$ layer is designed to have the same kernel size, dilation, and stride as the shared convolution layer such that they have the same receptive field. In addition, batch normalization is applied at the end of convolutional layer for a scale adjustment.

\begin{figure*}[t]
\centering
\includegraphics[width=0.95\textwidth]{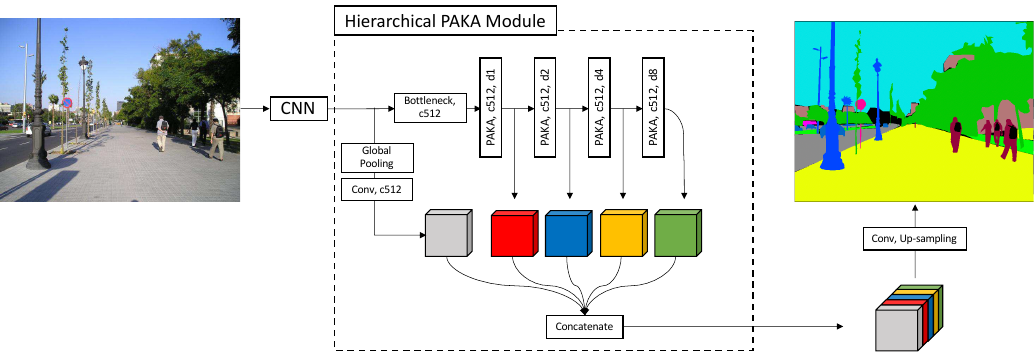}
\caption{An architecture of HPM. We apply the bottleneck layer to reduce the channel of input feature maps before feeding into the first PAKA layer. The module contains several PAKA layers with different dilation rates. We indicate PAKA layers' output channel with c and dilation rate with d. We concatenate the input and output features of each PAKA layer to utilize various receptive fields.}
\label{fig:net}
\end{figure*}

\subsection{Hierarchical PAKA Module}

With PAKA, we propose our hierarchical PAKA module, or HPM, as illustrated in Fig.~\ref{fig:net}. On the top of the module, we apply a $1\times1$ convolutional layer with BN and ReLU to squeeze the number of channels before feeding into the convolutional layers with PAKA. The HPM employs a series of dilated convolutional layers with PAKA to deal with the hierarchical pyramid architecture in fixed size feature maps. Using dilated convolutional layers with increasing dilation rates, the module covers diverse receptive fields as similar to the hierarchical pyramid architecture. In the last of HPM, we concatenate the Globally pooling input and every output feature maps from PAKA with different dilation rates. 

\subsection{Understanding and Analysis}
\label{sec:anal}

Several existing convolution operations perform position-specific modifications as explained in Section~\ref{sec:related}. In DFN~\cite{dfn}, since an auxiliary network generates filters for every offset and channel, a large number of parameters are needed. In addition, DFN requires an elaborate architecture design because all position-specific filter weights have to be predicted without sharing. Unlike DFN, PAKA allows efficient learning by position-specific attentions and shared weights. Liu~\etal~\cite{spn} and Cheng~\etal~\cite{cspn} propose spatial propagation networks by learning affinity. Although they utilize the directivity to propagate the information, they target the guide learning or refinement to learn affinity matrices. On the contrary, PAKA can be self-directed because it only specifies the direction to focus on each pixel.

\begin{figure}[t]
\centering
\includegraphics[width=0.8\linewidth]{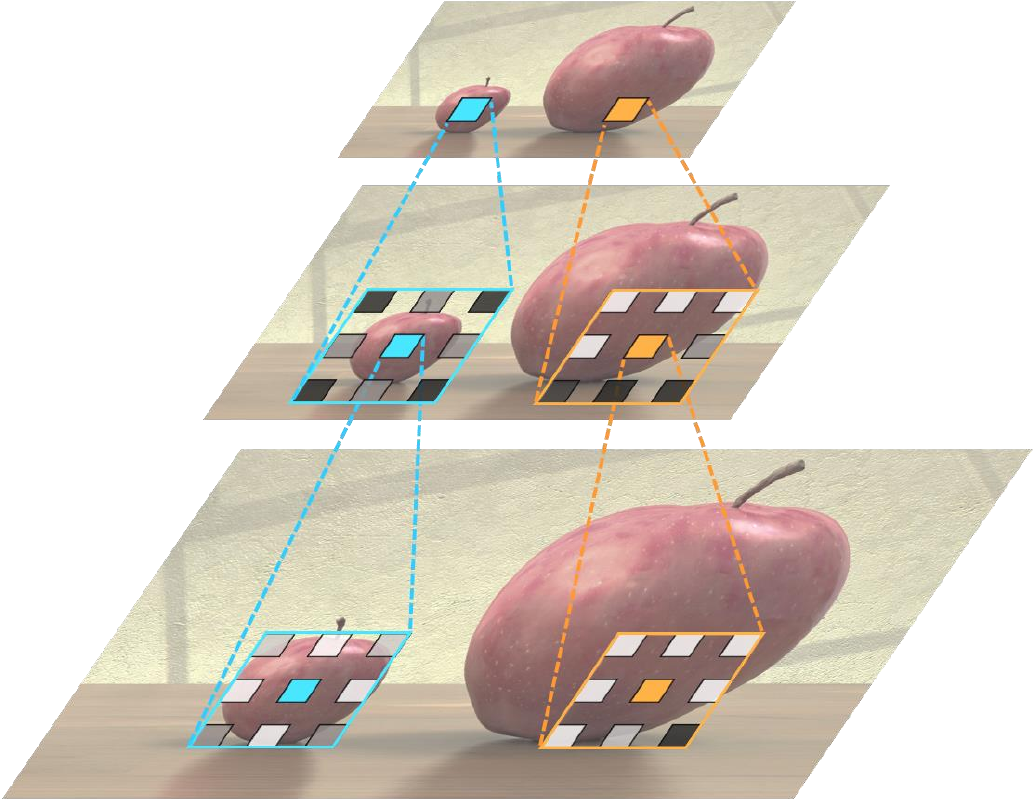}
\caption{Toy example for multiple convolutional layers with PAKA on different locations. With PAKA, the convolution propagates the information from different directions depending on the local contents.}
\label{fig:toy}
\end{figure}

\begin{figure*}
    \centering
    \subfigure[]
    {
        \includegraphics[width=0.17\textwidth]{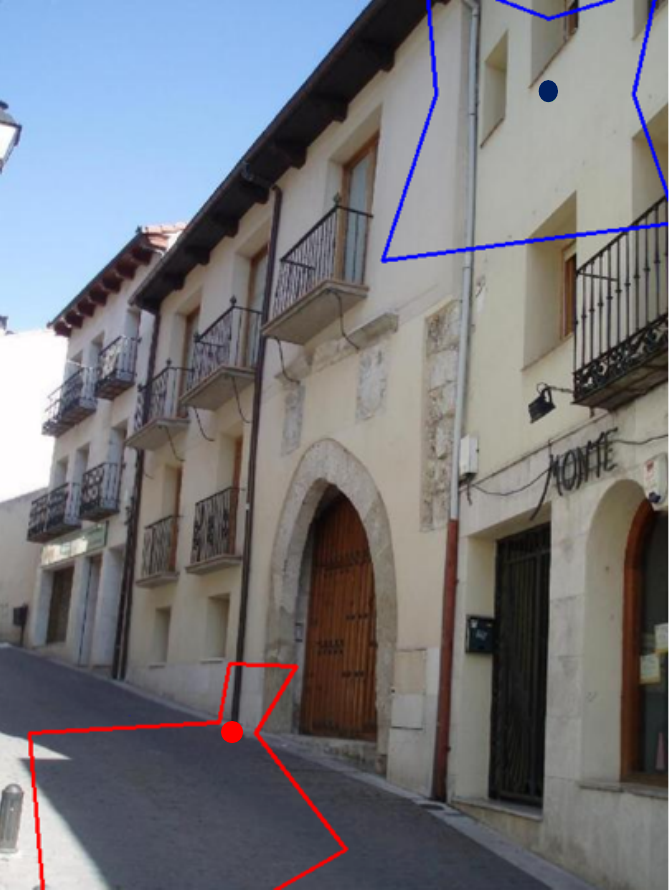}
        \label{fig:pic1}
    }
    \subfigure[]
    {
        \includegraphics[width=0.155\textwidth]{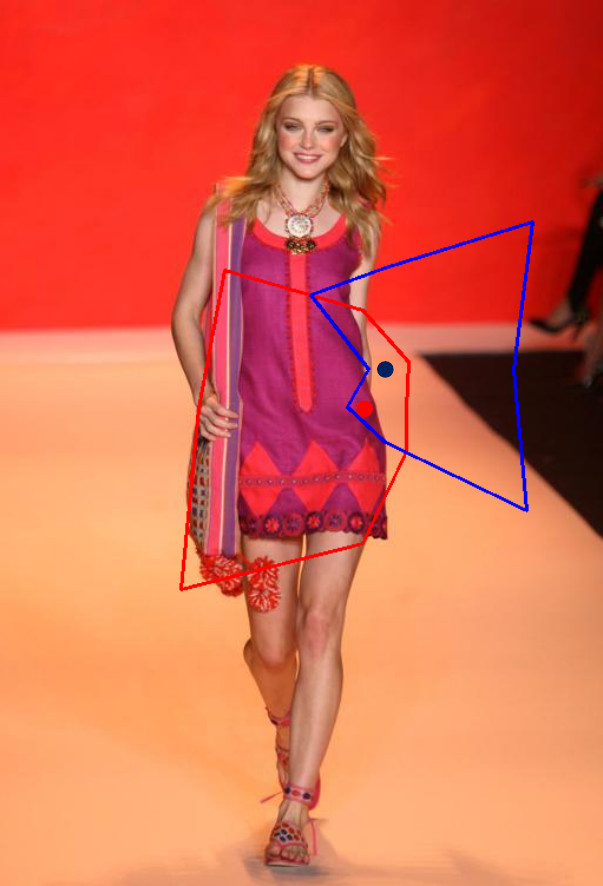}
         \label{fig:pic2}
    }
    \subfigure[]
    {
        \includegraphics[width=0.3\textwidth]{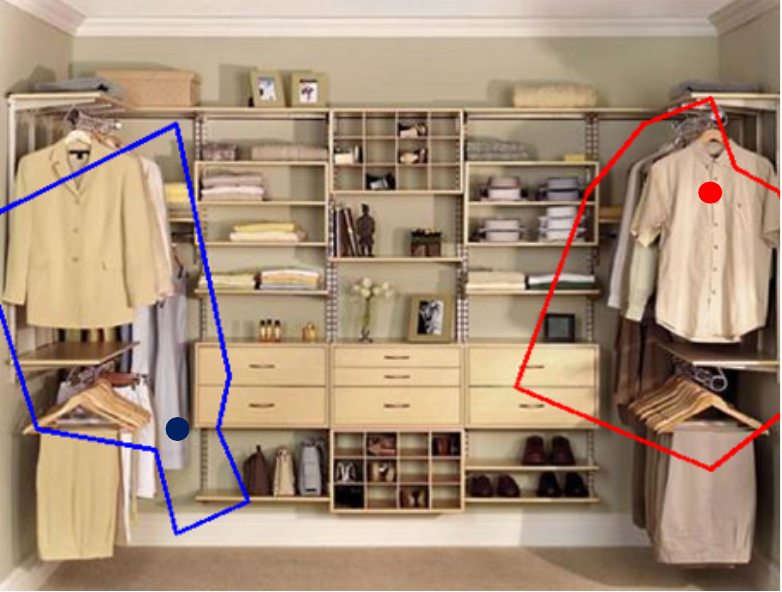}
         \label{fig:pic3}
    }
    \subfigure[]
    {
        \includegraphics[width=0.3\textwidth]{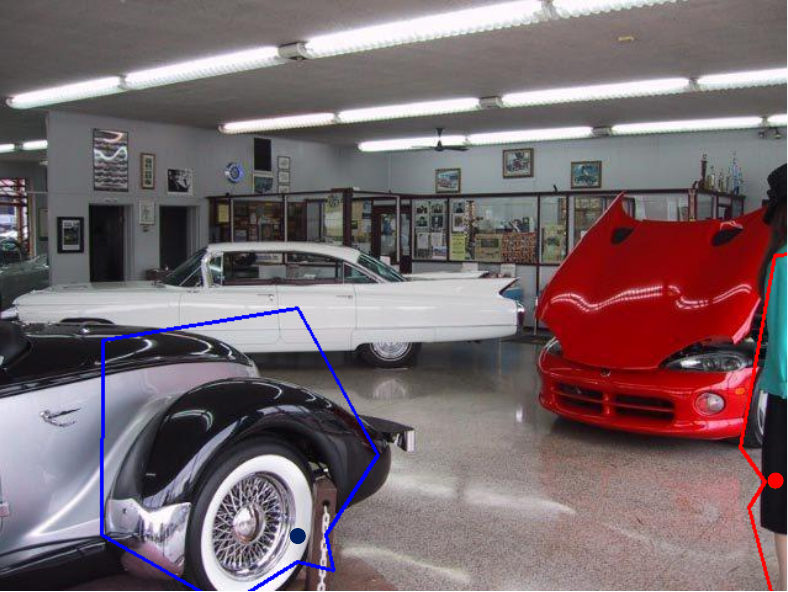}
         \label{fig:pic4}
    }
    \caption{Visualization of the propagational fields corresponding to the marked pixels. Each sub-figure shows the propagational fields of the pixels on different contents.}
    \label{fig:results}
\end{figure*}

\vspace{0.3cm}
\begin{figure*}
    \centering
    \subfigure[]
    {
        \includegraphics[width=0.25\textwidth]{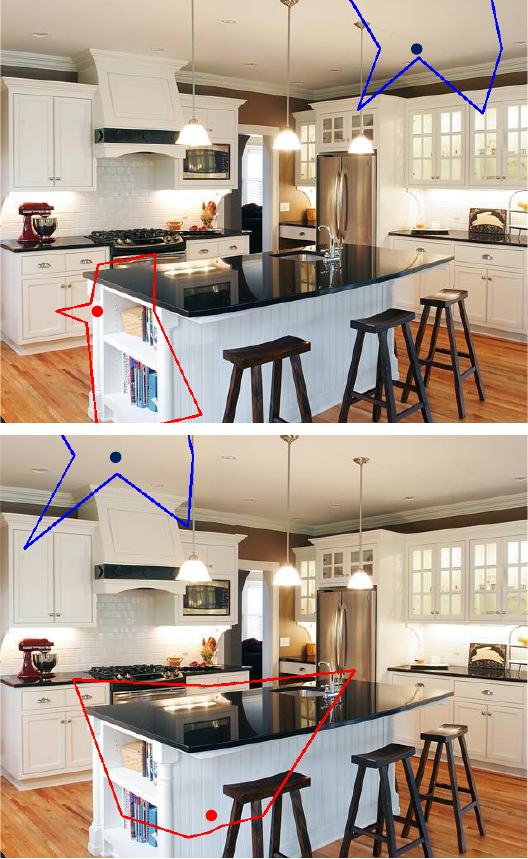}
        \label{fig:pic5}
    }
    \subfigure[]
    {
        \includegraphics[width=0.3\textwidth]{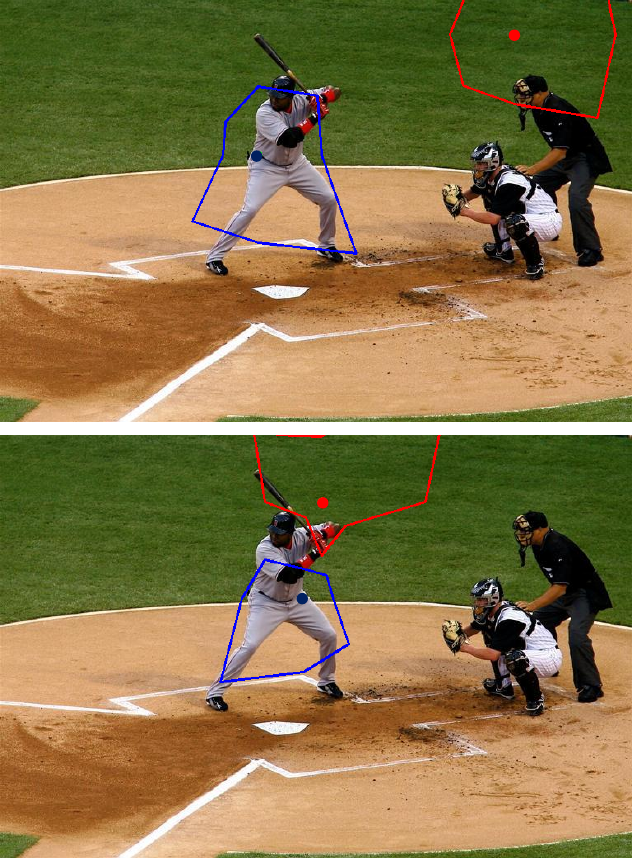}
         \label{fig:pic6}
    }
    \subfigure[]
    {
        \includegraphics[width=0.367\textwidth]{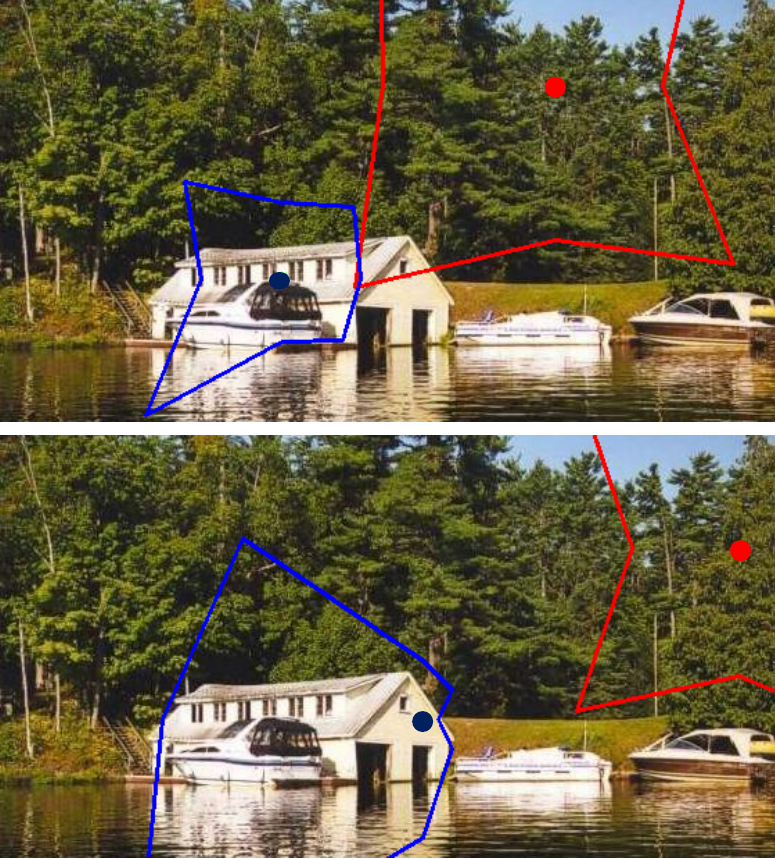}
         \label{fig:pic7}
    }
    \caption{Visualization of the propagational fields corresponding to the marked pixels. Each sub-figure shows the changes of the propagational fields when pixel location is changed even in the same object.}
    \label{fig:results2}
\end{figure*}

The deformable convolution~\cite{deformable, deformable2} employs position-specific modification by altering the grid of convolutional kernel to apply different sizes of receptive fields on different targets. It augments the spatial sampling location with the learnable offsets and modulation scalars. However, since the offset vectors have a high degree of freedom, it is very challenging to reach an optimal solution. In addition, the deformable convolution cannot explicitly consider the channel-wise attention because it only modifies the spatial sampling location. Fig.~\ref{fig:toy} shows how PAKA applies different effective receptive fields with kernel attention. We adopt PAKA in a hierarchical pyramid architecture to adjust effective receptive fields by regulating kernel attention in different layers. Consequently, various effective patch sizes can be utilized even with adoption of the fixed kernel size. As a result, PAKA drives the network to take information from different effective receptive fields depending on the contents.

Figs.~\ref{fig:results}~and~\ref{fig:results2} visualize the modulated receptive fields obtained by PAKA, named propagational field. Here, the propagational field is visualized by adding 8-neighborhood sampling offsets multiplied by the directional modulation from every PAKA layer to indicate which directions received high attention. Fig.~\ref{fig:results} illustrates the propagational fields of two pixels on different objects. As can be seen especially from Fig.~\ref{fig:pic2}, the propagational fields of the adjacent pixels on the human and the background are clearly distinguished. In addition, Fig.~\ref{fig:results2} demonstrates that even pixels corresponding to the same contents inherit information from different propagational fields depending on their locations. 

\section{Experimental Results}

The proposed method is evaluated on the ADE20K~\cite{ade} dataset for the semantic segmentation task. The ADE20K dataset contains very challenging 150 classes including 35 stuff classes and 115 discrete object classes. The dataset is divided into 20,210 images for training, 2,000 and 3,352 images for validation and testing, respectively. For evaluation, \textit{class-wise intersection over union}~(mIoU) and \textit{pixel-wise accuracy}~(PixAcc) are used.

\subsection{Implementation Details}

% \textbf{Cityscapes}
% The dataset consists of road scene images for scene parsing with 19 classes. It contains 2975, 500, 1525 images for training, validation, and testing, respectively. Each image has $2048\times2048$ resolution.

As a backbone network for feature extraction, we use a pretrained ResNet~\cite{resnet} model with the dilated network strategy~\cite{deeplab1, dilation}. The size of the feature maps is 1/8 of the size of the input image. We adopt convolutional layers and bilinear up-sampling layer to generate the final prediction map. We apply the cross-entropy loss to train the proposed network. Following the previous researches~\cite{pspnet}, we integrate the auxiliary loss in stage 4 of the ResNet backbone.

For the training, we normalize the input in the range~$[-1, 1]$. We also apply data augmentation including random horizontal flipping, random Gaussian filtering, and scaling with random factors of $[0.5, 2.0]$ to avoid overfitting. Last, we randomly crop the input image into the fixed size of $256 \times 256$ pixels. The pioneering researches~\cite{pspnet, cem} mention that the larger the crop size, the better the semantic segmentation performance. However, we use the same patch size of $256 \times 256$ for all compared methods considering our hardware resource. Although it is smaller than the size used in the original papers, whole conditions are unified for fair validation of the information aggregation modules.

During the training phase, we employ the stochastic gradient descent algorithm with a poly learning rate policy, $\gamma = \gamma_0 \times (1-\frac{N_{iter}}{N_{total}})^p$, where $N_{iter}$ and $N_{total}$ represent the current iteration number and total iteration number, respectively, and $p=0.9$. We set the initial learning rate $\gamma_0$ as 0.01, momentum as 0.9, weight decay as 0.0001, and batch size as 16. We train the model for 150K iterations. 

\subsection{Evaluation for Semantic Segmentation}

\begin{table}[t]
\footnotesize
\caption{Segmentation results on the ADE20K validation set in comparison with different information aggregation modules.}
\label{table:module}
\begin{center}
\begin{tabular}{l|c|c}
\hline
model & mIOU & pixAcc\\
\hline\hline
ResNet-50 (Baseline) & 36.28 & 76.84 \\

ResNet-50 + PSP & 38.34 & 77.79 \\

ResNet-50 + Self-Attention & 39.16 & 78.58 \\

ResNet-50 + ASPP & 40.37 & 79.39 \\

% ResNet-50 + CEM & 36.-- & 77.-- \\

ResNet-50 + HPM (Proposed) & \textbf{41.15} & \textbf{79.70} \\
\hline\hline
ResNet-101 (Baseline) & 39.03 & 78.60 \\

ResNet-101 + PSP & 40.12 & 79.55 \\

ResNet-101 + Self-Attention & 41.84 & 80.08 \\

ResNet-101 + ASPP & 42.08 & 80.07 \\

ResNet-101 + HPM (Proposed) & \textbf{42.21} & \textbf{80.38} \\

\hline
\end{tabular}
\end{center}

\vspace{0.5cm}
\footnotesize
\caption{Ablation study demonstrating the effects of channel modulation and directional modulation.}
\label{table:ablation2}
\begin{center}
\begin{tabular}{cccc}
\hline
Channel & Directional & mIOU & pixAcc\\
\hline\hline
& & 39.85 & 78.43 \\
\checkmark & & 40.41 & 79.68\\
& \checkmark & 40.85 & 79.65 \\
\checkmark & \checkmark & \textbf{41.15} & \textbf{79.70} \\
\hline
\end{tabular}
\end{center}

\vspace{0.5cm}
\footnotesize
\caption{Ablation study replacing standard convolution \\ with PAKA for ASPP and replacing PAKA with standard convolution for HPM.}
\label{table:ablation}
\begin{center}
\begin{tabular}{c|cccc}
\hline
Backbone & \multicolumn{4}{c}{ResNet-50}\\
\hline
 Aggregation module& \multicolumn{2}{c}{ASPP} & \multicolumn{2}{c}{HPM} \\
 \hline
 PAKA & & \checkmark & & \checkmark\\
\hline\hline
\#~parameters & 56.0M & 90.7M & 36.1M & 40.8M\\
average runtime & 138ms & 187ms & 124ms & 136ms\\
mIOU & 40.37 & \underline{40.71} & 39.85  & \textbf{41.15}  \\
\hline
\end{tabular}
\vspace{-0.3cm}
\end{center}
\end{table}

\textbf{Ablation Study for HPM}
To demonstrate the superiority of our PAKA and HPM, we conduct the experiments with several different settings containing the conventional information aggregation modules, which are PSP~\cite{pspnet}, ASPP~\cite{deeplab2} and, Self-Attention~\cite{self}, as shown in Table~\ref{table:module}. 

\begin{figure}[t]
\centering
\includegraphics[width=\linewidth]{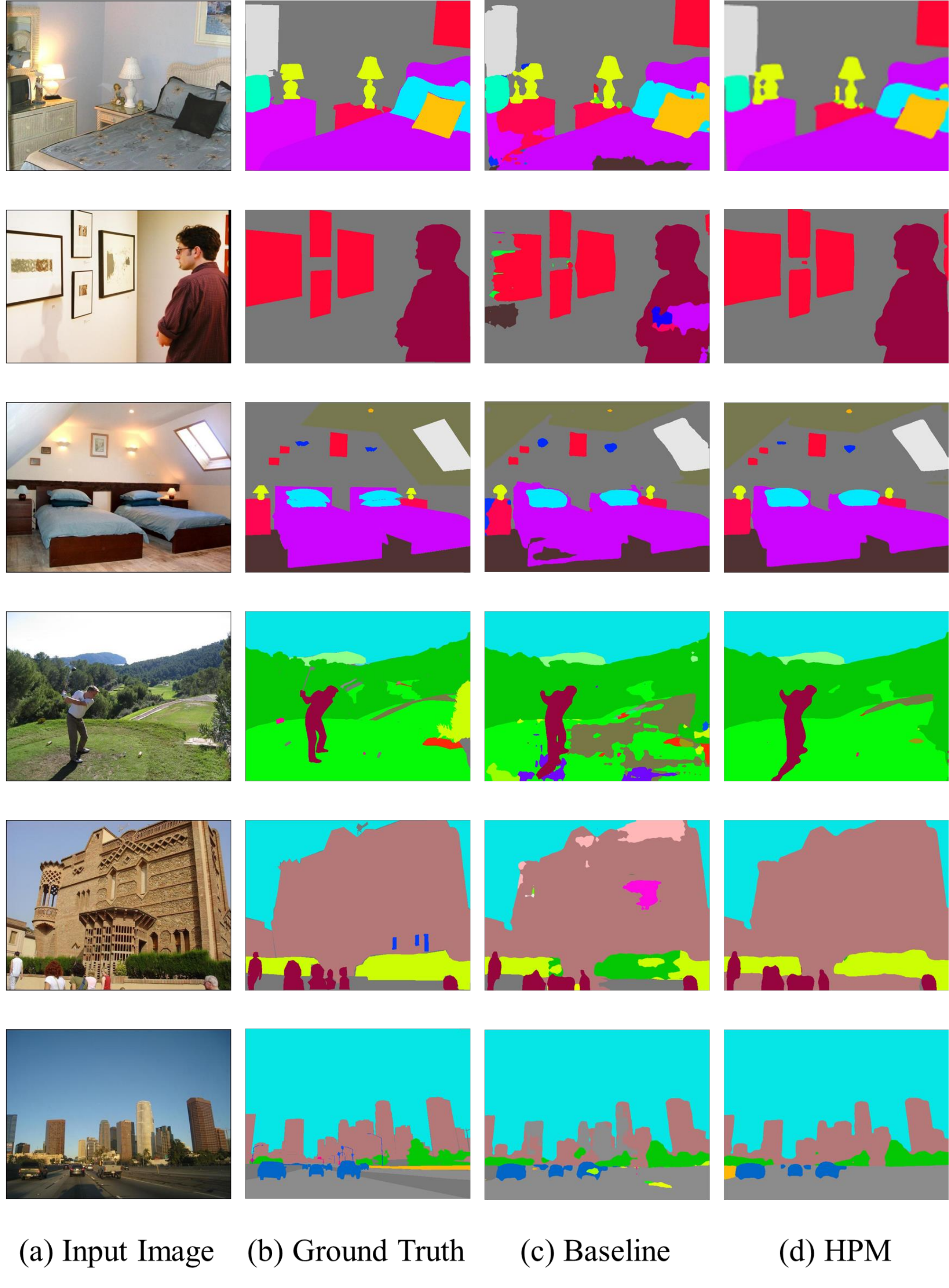}
\caption{Qualitative comparisons with the baseline and HPM on the ADE20K validation dataset. Propagation of the related context information by PAKA helps to understand the scenes, resulting in a significantly performance improvement over the baseline.}
\label{fig:reseg}
\end{figure}

First, we evaluate performance of our baseline model. The baseline consists of ResNet for the backbone network and the last convolutional layers followed by a up-sampling layer. It results in 36.28\% in terms of mIoU and 76.84\% in terms of PixAcc. We adopt this model as our baseline to compare the performance of the information aggregation modules. We list our evaluation results of different information aggregation modules in Table~\ref{table:module}. Note that we reimplement the conventional methods using the same condition for fair comparison. As can be seen in the table, the proposed method shows the best result in terms of both mIoU and PixAcc. With our HPM, the baseline model is significantly improved by 4.87\% and 2.86\% in terms of mIoU and PixAcc, respectively. Fig.~\ref{fig:reseg} shows several semantic segmentation results obtained by the baseline and HPM, demonstrating the effectiveness of information aggregation by HPM. 

In order to demonstrate the validity of the proposed modulations, we conducted ablation experiments by eliminating each modulation from PAKA. Table~\ref{table:ablation2} shows that the PAKA without any modulation exhibits inferior performance compared with the original PAKA and both modulations are essential for the performance improvement. To support the effectiveness of PAKA and the necessity of HPM, we have conducted ablation studies by switching the standard convolution and PAKA in ASPP and HPM. As shown in Table~\ref{table:ablation}, simple replacement of the standard convolution by PAKA improves the performance of ASPP. In addition, it shows that PAKA is more effective with HPM compared to ASPP. We consider this is because a cascade structure with PAKA can utilize diverse propagational fields as illustrated in Fig.~\ref{fig:toy}. However, ASPP adopts a parallel structure that cannot fully take advantage of PAKA. Therefore, we designed HPM including a cascade structure with PAKA to fully exploit its effectiveness. Moreover, to show the computational cost of PAKA, we present the number of parameters and the runtime of ASPP and HPM in Table~\ref{table:ablation}. ASPP requires much more parameters than HPM for the adoption of PAKA. This is because ASPP adopts a parallel structure with 2,048 input channels for every layer but HPM adopts a cascade structure with 512 input channels. As a result, we verify that PAKA improves not only the conventional aggregation module such as ASPP but also HPM more significantly with much less parameter increment.

We also evaluate the combined model of HPM and the conventional modules. As mentioned in Section~\ref{sec:anal}, HPM learns direction to propagate the context information. Since HPM plays a different role compared to other modules, combining HPM with the other existing modules can yield further improvement. As indicated in Table~\ref{table:delta}, the combined module produces better results than a single module.

\textbf{Comparison with state-of-the-art methods}
We compare the proposed method with other state-of-the-art methods to demonstrate effectiveness of the proposed method. For this experiment, we use the same patch size as the conventional methods~($480 \times 480$) for training and utilize the combined model of ASPP and HPM. Similar to the compared methods~\cite{pspnet, deeplab2, acnet, cpn}, we average the predictions from multiple scaled and flipped inputs to further improve the performance. We use the scale factors of $\{0.5, 0.75, 1.0, 1.25, 1.5\}$ for the multi-scale testing strategy. As shown in Table~\ref{table:seg}, the proposed method outperforms recent state-of-the-art methods although our proposed method is trained with simple up-sampling layers for image reconstruction. % Note that, we do not intend to seek an optimal network architecture for the semantic segmentation task but to validate our novel convolutional operation.

\vspace{0.3cm}
\begin{table}[t]
\centering
\caption{Performance improvements by combining HPM with the conventional modules. The evaluation metric is mIoU(\%). $\Delta$ indicates the absolute difference.}
\label{table:delta}
\begin{tabular}{ c | ccc}
\hline 
 & PSP & Self-Attention & ASPP \\
\hline\hline
w/o HPM & 38.34 & 39.16 & 40.37 \\

w/ HPM & 41.48 & 41.02 & 41.47 \\

\hline
$\Delta$ & $\uparrow$3.14 & $\uparrow$1.86 & $\uparrow$1.10 \\

\hline
\end{tabular}
\end{table}

\begin{table}[t]
\centering
\caption{Quantitative evaluation on the ADE20K validation set in comparison with state-of-the-art methods.}
\label{table:seg}
\begin{tabular}{l|c|c|c}
\hline
Method & Backbone & mIOU & pixAcc\\
\hline\hline
EncNet~\cite{cem} & ResNet-50 & 41.11 & 79.73 \\

PSPNet~\cite{pspnet} & ResNet-50 & 42.78 & 80.76\\

ACNet~\cite{acnet} & ResNet-50 & 43.01 & 81.01 \\

CFNet~\cite{cfnet} & ResNet-50 & 42.87 & - \\

CPN~\cite{cpn} & ResNet-50 & 44.46 & 81.38\\

Proposed & ResNet-50 & \textbf{44.75} & \textbf{81.61} \\
\hline\hline
PSPNet~\cite{pspnet} & ResNet-101 & 43.29 & 81.39 \\

PSANet~\cite{psanet} & ResNet-101 & 43.77 & 81.51\\

EncNet~\cite{cem} & ResNet-101 & 44.65 & 81.19 \\

CFNet~\cite{cfnet} & ResNet-101 & 44.89 & - \\

ANL~\cite{asymmetric} & ResNet-101 & 45.24 & -\\

OCR~\cite{ocr} & ResNet-101 & 45.28 & -\\

APCNet~\cite{apcnet} & ResNet-101 & 45.38 & -\\

Proposed & ResNet-101 & \textbf{45.42} & \textbf{81.88} \\

\hline
\end{tabular}
\end{table}

\section{Joint Depth Super-resolution with PAKA}
In order to demonstrate superiority over other content-adaptive convolution, \ie, pixel-adaptive convolution~(PAC), and generalizability of PAKA, as a case study, we apply PAKA for the color-guided depth map super-resolution task, which generates a high-resolution depth map with help of its corresponding high-resolution color image. Early studies adopt content-adaptive filtering techniques such as joint bilateral filtering~\cite{jointbilateral} and guided image filtering~\cite{guided}. The common characteristic of them is that they determine filter weights to up-sample the target~(\ie, low-resolution (LR) depth map) from the guide content~(\ie, HR color image). Consequently, the surrounding pixels that have higher content similarity to the center have greater influence in filtering. We notice that PAKA can be used as a learnable and generalized model of such conventional filtering methods. 

\subsection{Joint up-sampling layer with PAKA}
\begin{figure}[t]
\centering
\includegraphics[width=\linewidth]{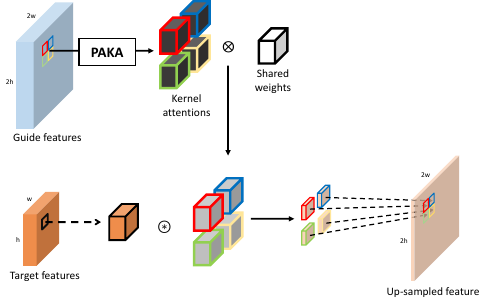}
\caption{Framework of the proposed joint up-sampling layer with PAKA for depth map super-resolution. The proposed joint up-sampling layer generate 4~sub-pixels from LR target feature maps for $\times2$ up-sampling.}
\label{fig:joint}
\end{figure}

To conduct the color-guided depth map super-resolution task, we design a joint up-sampling layer with PAKA. Fig.~\ref{fig:joint} illustrates the proposed joint up-sampling layer with PAKA. The proposed joint up-sampling layer receives low-resolution~(LR) target feature maps and high-resolution~(HR) guide feature maps as input. We obtain the kernel attentions from the guide feature maps by employing both channel modulation branch and directional modulation branch in the same manner as the original PAKA. The each kernel attention is multiplied with the shared weights to generate sub-pixels of the upsampled feature maps. With the kernel attention from PAKA, the proposed joint up-sampling layer is driven to concentrate more on important directions~(\eg edge) to inherit the information from the guide features.

\subsection{Implementation Details}
\begin{figure}[t]
\centering
\includegraphics[width=\linewidth]{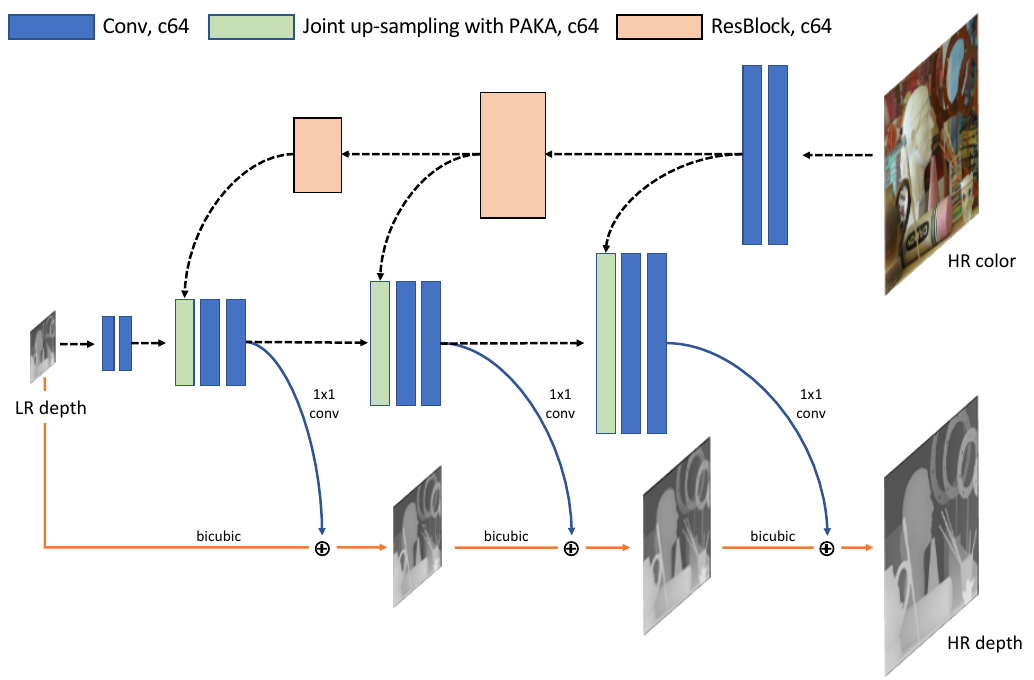}
\caption{Architecture of the proposed color guided depth-map super-resolution network for $\times 8$ up-sampling. We apply our joint up-sampling layer in Fig.~\ref{fig:joint} to the architecture motivated from the MSG-Net~\cite{msg}.}
\label{fig:dsr}
\end{figure}

We build the network architecture motivated by the pioneering method~\cite{msg} of joint depth map super-resolution, which is simple yet efficient. We provide the details of the network in Fig.~\ref{fig:dsr}. Following the common training procedure~\cite{msg, hdsr}, we use MPI Sintel depth dataset~\cite{sintel} and Middlebury dataset~\cite{middle}~(including 2001, 2006 and 2014 datasets). In the training phase, we crop the images into $128\times 128$ patches with overlapping for all scaling factors~(\ie $\times8$ and $\times16$) to reduce the training time.

\subsection{Evaluation for depth map super-resolution} 
\begin{figure*}[t]
\centering
\includegraphics[width=\linewidth]{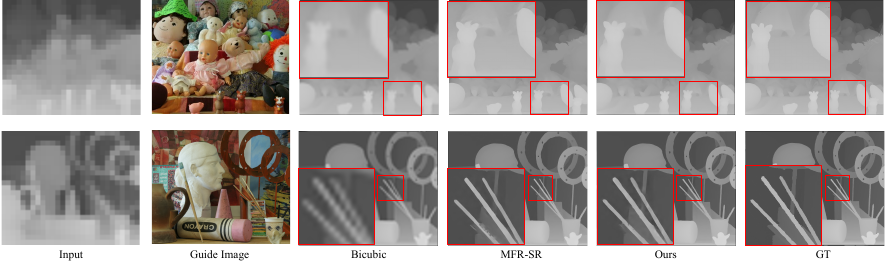}
\caption{Visual comparison results at $\times 16$ up-sampling for ``Dolls'' and ``Art'' in the Middlebury dataset. The regions inside the red boxes are magnified for comparison.}
\label{fig:redsr}
\end{figure*}

Table~\ref{table:dsr} shows the numeric results of the proposed and conventional methods in case of 8~times and 16~times super-resolution, respectively. We compare the performance in terms of \textit{root mean squared error}~(RMSE) and \textit{peak signal-to-noise ratio}~(PSNR). As can be seen in the table, our proposed joint up-sampling network with PAKA outperforms the conventional state-of-the-art methods~\cite{hdsr, zuo2019multi, pac}. For further study, we test the conventional methods in conjunction with PAKA. As can be seen in Table~\ref{table:replace}, PAKA contributes to further performance improvements of the conventional models. Fig.~\ref{fig:redsr} shows visual comparisons of the state-of-the-art methods and the proposed method. The proposed method generates more sharp object boundaries than other methods. It indicates that PAKA can help to learn accurate directivity to up-sample the features than the conventional methods.

\vspace{0.3cm}

\begin{table*}
    \centering
    \caption{Quantitative comparison on the Middlebury dataset for $\times 8$ (top) and $\times 16$ (bottom) scaling factors in terms of RMSE/PSNR (dB).}
    \label{table:dsr}
    \subtable
    {
        \centering
        \begin{tabular}{c|c|c|c|c|c|c|c}
        \hline
        Method & Art & Books & Dolls & Laundry & Moebius & Reindeer  & Average \\
        \hline\hline
        Bicubic & 5.47 / 33.37 & 2.34 / 40.75 & 1.87 / 42.69 & 3.42 / 37.45 & 1.96 / 42.29 & 4.01 / 36.07 & 3.18 / 38.01 \\

        TGV~\cite{tgv} & 7.02 / 31.20 & 2.08 / 41.77 & 2.05 / 41.90 & 3.92 / 36.27 & 2.41 / 40.49 & 4.29 / 35.48 & 3.63 / 37.85 \\

        SRCNN~\cite{srcnn} & 4.75 / 34.60 & 2.15 / 41.48 & 1.92 / 42.46 & 3.45 / 37.37 & 2.00 / 42.11 & 3.87 / 36/38 & 3.02 / 38.52 \\

        MSG-Net~\cite{msg} & 2.63 / 39.73 & 1.16 / 46.87 &  1.67 / 43.79 & 1.67 / 43.69 & 1.21 / 46.51 & 1.97 / 42.24 &  1.64 / 43.81 \\

        Depth-SR~\cite{hdsr} & 2.39 / 40.58 & 1.04 / 47.75 &  1.22 / 46.37 & 1.43 / 45.03 & 1.10 / 47.27 & 1.78 / 43.14 & 1.43 / 45.02 \\

        MFR-SR~\cite{zuo2019multi} & \underline{2.29 / 40.94} & \underline{1.02 / 47.98} & \underline{1.22 / 46.38} & 1.41 / 45.15 & 1.10 / 47.27 & 1.68 / 43.61 & \underline{1.40 / 45.22} \\

        PAC~\cite{pac} & 2.36 / 40.67 & 1.05 / 47.74 & 1.24 / 46.23 & \underline{1.40 / 45.22} & \underline{1.10 / 47.32} & \underline{1.66 / 43.73} & 1.41 / 45.15 \\

        Proposed & \textbf{2.26 / 41.04} & \textbf{0.83} / \textbf{49.71} & \textbf{1.12} / \textbf{47.16} & \textbf{1.30} / \textbf{45.83} & \textbf{0.89} / \textbf{49.15} & \textbf{1.61} / \textbf{44.01} & \textbf{1.26} / \textbf{46.15} \\
        \hline
        \end{tabular}
    }
    \subtable
    {
        \centering
        \begin{tabular}{c|c|c|c|c|c|c|c}
        \hline
        Method & Art & Books & Dolls & Laundry & Moebius & Reindeer  & Average \\
        \hline\hline
        Bicubic & 8.17 / 29.89 & 3.34 / 37.65 & 2.64 / 39.70 & 5.07 / 34.04 & 2.85 / 39.03 & 5.86 / 32.77 & 4.66 / 34.77 \\

        TGV~\cite{tgv} & 12.08 / 26.49 & 4.89 / 34.34 & 4.44 / 35.18 & 8.01 / 30.06 & 5.41 / 33.47 & 9.05 / 29.00 & 7.31 / 31.42 \\

        SRCNN~\cite{srcnn} & 7.80 / 30.29 & 3.24 / 37.91 & 2.61 / 39.80 & 5.04 / 34.08 & 2.82 / 39.13 & 5.63 / 33.12 & 4.52 / 35.02 \\

        MSG-Net~\cite{msg} & 4.25 / 35.57 & 1.85 / 42.81 &  1.77 / 43.19 & 2.92 / 38.81 & 1.79 / 43.05 & 3.18 / 38.09 & 2.48 / 40.25 \\

        Depth-SR~\cite{hdsr} & 4.09 / 35.89 & 1.65 / 43.78 & 1.68 / 43.63 & 2.31 / 40.86 & 1.66 / 43.71 & 2.68 / 39.58 & 2.21 / 41.24 \\

        MFR-SR~\cite{zuo2019multi} & \textbf{3.55 / 37.13} & \underline{1.60 / 44.06} & \underline{1.62 / 43.95} & \underline{2.18 / 41.35} & 1.58 / 44.18 & \textbf{2.35 / 40.70} & \underline{2.05 / 41.90} \\

        PAC~\cite{pac} & 3.75 / 36.66 & 1.64 / 43.86 & 1.62 / 43.92 & 2.28 / 40.96 & \underline{1.53 / 44.42} & 2.59 / 39.87 & 2.12 / 41.62 \\

        Proposed & \underline{3.58 / 37.04} & \textbf{1.44} / \textbf{44.98} & \textbf{1.54} / \textbf{44.36} & \textbf{2.17} / \textbf{41.39} & \textbf{1.44} / \textbf{44.97} & \underline{2.43 / 40.41} & \textbf{1.98} / \textbf{42.19} \\
        \hline
        \end{tabular}
    }
\end{table*}

\begin{table}[t]
\centering
\caption{Direct replacement results with the conventional methods on the depth map super resolution in terms of average PSNR~(dB).}
\label{table:replace}
\begin{tabular}{l|c|c}
\hline
Method & PSNR~($\times$8) & PSNR~($\times$16) \\
\hline\hline

MSG-Net~\cite{msg} & 43.81 & 40.25\\

MSG-Net w/ PAKA & 44.58 & 41.66 \\

PAC-Net~\cite{pac} & 45.15 & 41.61 \\

PAC-Net w/ PAKA & 45.31 & 41.75 \\

Proposed & \textbf{45.78} & \textbf{41.91} \\
\hline
\end{tabular}
\end{table}

\section{Conclusion}

In this paper, we propose a novel convolutional operation called PAKA that learns the directivity to effectively propagate information. PAKA emphasizes or suppresses information from different directions with two decomposed branches, which predict channel and directional modulations. We validate superiority of PAKA with visualization of its propagational fields and the extensive experiments. We also demonstrate that PAKA is applicable in many vision tasks including semantic segmentation and joint depth map super-resolution. Since PAKA can directly replace the standard convolutional layers, we think that PAKA has a notable potential to leverage various CNN-based tasks.

\bibliographystyle{IEEEtran}

\bibliography{bib_cc.bib}

% Generated by IEEEtran.bst, version: 1.14 (2015/08/26)
\begin{thebibliography}{10}
\providecommand{\url}[1]{#1}
\csname url@samestyle\endcsname
\providecommand{\newblock}{\relax}
\providecommand{\bibinfo}[2]{#2}
\providecommand{\BIBentrySTDinterwordspacing}{\spaceskip=0pt\relax}
\providecommand{\BIBentryALTinterwordstretchfactor}{4}
\providecommand{\BIBentryALTinterwordspacing}{\spaceskip=\fontdimen2\font plus
\BIBentryALTinterwordstretchfactor\fontdimen3\font minus
  \fontdimen4\font\relax}
\providecommand{\BIBforeignlanguage}[2]{{%
\expandafter\ifx\csname l@#1\endcsname\relax
\typeout{** WARNING: IEEEtran.bst: No hyphenation pattern has been}%
\typeout{** loaded for the language `#1'. Using the pattern for}%
\typeout{** the default language instead.}%
\else
\language=\csname l@#1\endcsname
\fi
#2}}
\providecommand{\BIBdecl}{\relax}
\BIBdecl

\bibitem{fastrcnn}
R.~Girshick, ``Fast {R-CNN},'' in \emph{Proceedings of the IEEE International
  Conference on Computer Vision}, 2015, pp. 1440--1448.

\bibitem{deformable}
J.~Dai, H.~Qi, Y.~Xiong, Y.~Li, G.~Zhang, H.~Hu, and Y.~Wei, ``Deformable
  convolutional networks,'' in \emph{Proceedings of the IEEE International
  Conference on Computer Vision}, 2017, pp. 764--773.

\bibitem{cornernet}
H.~Law and J.~Deng, ``Cornernet: Detecting objects as paired keypoints,'' in
  \emph{Proceedings of the European Conference on Computer Vision}, 2018, pp.
  734--750.

\bibitem{Cao_2020_CVPR}
Y.~Cao, K.~Chen, C.~C. Loy, and D.~Lin, ``Prime sample attention in object
  detection,'' in \emph{Proceedings of the IEEE Conference on Computer Vision
  and Pattern Recognition}, June 2020.

\bibitem{Tan_2020_CVPR}
M.~Tan, R.~Pang, and Q.~V. Le, ``{EfficientDet}: Scalable and efficient object
  detection,'' in \emph{Proceedings of the IEEE Conference on Computer Vision
  and Pattern Recognition}, June 2020.

\bibitem{ji2020casnet}
Y.~Ji, H.~Zhang, Z.~Jie, L.~Ma, and Q.~J. Wu, ``Casnet: A cross-attention
  siamese network for video salient object detection,'' \emph{IEEE transactions
  on neural networks and learning systems}, vol.~32, no.~6, pp. 2676--2690,
  2020.

\bibitem{vgg}
K.~Simonyan and A.~Zisserman, ``Very deep convolutional networks for
  large-scale image recognition,'' \emph{arXiv preprint arXiv:1409.1556}, 2014.

\bibitem{resnet}
K.~He, X.~Zhang, S.~Ren, and J.~Sun, ``Deep residual learning for image
  recognition,'' in \emph{Proceedings of the IEEE Conference on Computer Vision
  and Pattern Recognition}, 2016, pp. 770--778.

\bibitem{inception}
C.~Szegedy, W.~Liu, Y.~Jia, P.~Sermanet, S.~Reed, D.~Anguelov, D.~Erhan,
  V.~Vanhoucke, and A.~Rabinovich, ``Going deeper with convolutions,'' in
  \emph{Proceedings of the IEEE Conference on Computer Vision and Pattern
  Recognition}, 2015, pp. 1--9.

\bibitem{xie2020self}
Q.~Xie, M.-T. Luong, E.~Hovy, and Q.~V. Le, ``Self-training with noisy student
  improves imagenet classification,'' in \emph{Proceedings of the IEEE
  Conference on Computer Vision and Pattern Recognition}, 2020, pp.
  10\,687--10\,698.

\bibitem{Zhang_2020_CVPR}
C.~Zhang, Y.~Cai, G.~Lin, and C.~Shen, ``{DeepEMD}: Few-shot image
  classification with differentiable earth mover's distance and structured
  classifiers,'' in \emph{Proceedings of the IEEE Conference on Computer Vision
  and Pattern Recognition}, June 2020.

\bibitem{pepsi}
M.-c. Sagong, Y.-g. Shin, S.-w. Kim, S.~Park, and S.-j. Ko, ``{PEPSI}: Fast
  image inpainting with parallel decoding network,'' in \emph{Proceedings of
  the IEEE Conference on Computer Vision and Pattern Recognition}, 2019, pp.
  11\,360--11\,368.

\bibitem{srcnn}
C.~Dong, C.~C. Loy, K.~He, and X.~Tang, ``Image super-resolution using deep
  convolutional networks,'' \emph{IEEE Transactions on Pattern Analysis and
  Machine Intelligence}, vol.~38, no.~2, pp. 295--307, 2015.

\bibitem{compression}
Y.-J. Yeo, Y.-G. Shin, M.-C. Sagong, S.-W. Kim, and S.-J. Ko, ``Simple yet
  effective way for improving the performance of lossy image compression,''
  \emph{IEEE Signal Processing Letters}, vol.~27, pp. 530--534, 2020.

\bibitem{Zamir_2020_CVPR}
S.~W. Zamir, A.~Arora, S.~Khan, M.~Hayat, F.~S. Khan, M.-H. Yang, and L.~Shao,
  ``{CycleISP}: Real image restoration via improved data synthesis,'' in
  \emph{Proceedings of the IEEE Conference on Computer Vision and Pattern
  Recognition}, June 2020.

\bibitem{Suin_2020_CVPR}
M.~Suin, K.~Purohit, and A.~N. Rajagopalan, ``Spatially-attentive
  patch-hierarchical network for adaptive motion deblurring,'' in
  \emph{Proceedings of the IEEE Conference on Computer Vision and Pattern
  Recognition}, June 2020.

\bibitem{gan}
I.~Goodfellow, J.~Pouget-Abadie, M.~Mirza, B.~Xu, D.~Warde-Farley, S.~Ozair,
  A.~Courville, and Y.~Bengio, ``Generative adversarial nets,'' in
  \emph{Proceedings of the Advances in Neural Information Processing Systems},
  2014, pp. 2672--2680.

\bibitem{style}
L.~A. Gatys, A.~S. Ecker, and M.~Bethge, ``Image style transfer using
  convolutional neural networks,'' in \emph{Proceedings of the IEEE Conference
  on Computer Vision and Pattern Recognition}, 2016, pp. 2414--2423.

\bibitem{Zhu_2020_CVPR}
P.~Zhu, R.~Abdal, Y.~Qin, and P.~Wonka, ``{SEAN}: Image synthesis with semantic
  region-adaptive normalization,'' in \emph{Proceedings of the IEEE Conference
  on Computer Vision and Pattern Recognition}, June 2020.

\bibitem{Lee_2020_CVPR}
C.-H. Lee, Z.~Liu, L.~Wu, and P.~Luo, ``{MaskGAN}: Towards diverse and
  interactive facial image manipulation,'' in \emph{Proceedings of the IEEE
  Conference on Computer Vision and Pattern Recognition}, June 2020.

\bibitem{pepsi++}
Y.-G. Shin, M.-C. Sagong, Y.-J. Yeo, S.-W. Kim, and S.-J. Ko, ``Pepsi++: Fast
  and lightweight network for image inpainting,'' \emph{IEEE Transactions on
  Neural Networks and Learning Systems}, vol.~32, no.~1, pp. 252--265, 2020.

\bibitem{dfn}
X.~Jia, B.~De~Brabandere, T.~Tuytelaars, and L.~V. Gool, ``Dynamic filter
  networks,'' in \emph{Proceedings of the Advances in Neural Information
  Processing Systems}, 2016, pp. 667--675.

\bibitem{pac}
H.~Su, V.~Jampani, D.~Sun, O.~Gallo, E.~Learned-Miller, and J.~Kautz,
  ``Pixel-adaptive convolutional neural networks,'' in \emph{Proceedings of the
  IEEE Conference on Computer Vision and Pattern Recognition}, 2019, pp.
  11\,166--11\,175.

\bibitem{deformable2}
X.~Zhu, H.~Hu, S.~Lin, and J.~Dai, ``Deformable convnets v2: More deformable,
  better results,'' in \emph{Proceedings of the IEEE Conference on Computer
  Vision and Pattern Recognition}, 2019, pp. 9308--9316.

\bibitem{bilateral}
C.~Tomasi and R.~Manduchi, ``Bilateral filtering for gray and color images,''
  in \emph{Proceedings of the IEEE International Conference on Computer
  Vision}, 1998, pp. 839--846.

\bibitem{gaussian}
V.~Aurich and J.~Weule, ``Non-linear {G}aussian filters performing edge
  preserving diffusion,'' in \emph{Mustererkennung 1995}.\hskip 1em plus 0.5em
  minus 0.4em\relax Springer, 1995, pp. 538--545.

\bibitem{guided}
K.~He, J.~Sun, and X.~Tang, ``Guided image filtering,'' \emph{IEEE Transactions
  on Pattern Analysis and Machine Intelligence}, vol.~35, no.~6, pp.
  1397--1409, 2012.

\bibitem{structure}
L.-C. Chen, A.~Schwing, A.~Yuille, and R.~Urtasun, ``Learning deep structured
  models,'' in \emph{Proceedings of the International Conference on Machine
  Learning}, 2015, pp. 1785--1794.

\bibitem{crf}
S.~Zheng, S.~Jayasumana, B.~Romera-Paredes, V.~Vineet, Z.~Su, D.~Du, C.~Huang,
  and P.~H. Torr, ``Conditional random fields as recurrent neural networks,''
  in \emph{Proceedings of the IEEE International Conference on Computer
  Vision}, 2015, pp. 1529--1537.

\bibitem{superpixel}
R.~Gadde, V.~Jampani, M.~Kiefel, D.~Kappler, and P.~V. Gehler, ``Superpixel
  convolutional networks using bilateral inceptions,'' in \emph{Proceedings of
  the European Conference on Computer Vision}, 2016, pp. 597--613.

\bibitem{end}
H.~Wu, S.~Zheng, J.~Zhang, and K.~Huang, ``Fast end-to-end trainable guided
  filter,'' in \emph{Proceedings of the IEEE Conference on Computer Vision and
  Pattern Recognition}, 2018, pp. 1838--1847.

\bibitem{sparse}
V.~Jampani, M.~Kiefel, and P.~V. Gehler, ``Learning sparse high dimensional
  filters: Image filtering, dense {CRF}s and bilateral neural networks,'' in
  \emph{Proceedings of the IEEE Conference on Computer Vision and Pattern
  Recognition}, 2016, pp. 4452--4461.

\bibitem{residualattention}
F.~Wang, M.~Jiang, C.~Qian, S.~Yang, C.~Li, H.~Zhang, X.~Wang, and X.~Tang,
  ``Residual attention network for image classification,'' in \emph{Proceedings
  of the IEEE Conference on Computer Vision and Pattern Recognition}, 2017, pp.
  3156--3164.

\bibitem{squeeze}
J.~Hu, L.~Shen, and G.~Sun, ``Squeeze-and-excitation networks,'' in
  \emph{Proceedings of the IEEE Conference on Computer Vision and Pattern
  Recognition}, 2018, pp. 7132--7141.

\bibitem{bam}
J.~Park, S.~Woo, J.-Y. Lee, and I.~S. Kweon, ``{BAM}: Bottleneck attention
  module,'' \emph{arXiv preprint arXiv:1807.06514}, 2018.

\bibitem{cbam}
S.~Woo, J.~Park, J.-Y. Lee, and I.~So~Kweon, ``{CBAM}: Convolutional block
  attention module,'' in \emph{Proceedings of the European Conference on
  Computer Vision}, 2018, pp. 3--19.

\bibitem{pspnet}
H.~Zhao, J.~Shi, X.~Qi, X.~Wang, and J.~Jia, ``Pyramid scene parsing network,''
  in \emph{Proceedings of the IEEE Conference on Computer Vision and Pattern
  Recognition}, 2017, pp. 2881--2890.

\bibitem{psp}
K.~He, X.~Zhang, S.~Ren, and J.~Sun, ``Spatial pyramid pooling in deep
  convolutional networks for visual recognition,'' \emph{IEEE Transactions on
  Pattern Analysis and Machine Intelligence}, vol.~37, no.~9, pp. 1904--1916,
  2015.

\bibitem{deeplab1}
L.-C. Chen, G.~Papandreou, I.~Kokkinos, K.~Murphy, and A.~L. Yuille, ``Semantic
  image segmentation with deep convolutional nets and fully connected {CRF}s,''
  \emph{arXiv preprint arXiv:1412.7062}, 2014.

\bibitem{deeplab2}
L.-C. Chen, G.~Papandreou, F.~Schroff, and H.~Adam, ``Rethinking atrous
  convolution for semantic image segmentation,'' \emph{arXiv preprint
  arXiv:1706.05587}, 2017.

\bibitem{deeplab3}
L.-C. Chen, Y.~Zhu, G.~Papandreou, F.~Schroff, and H.~Adam, ``Encoder-decoder
  with atrous separable convolution for semantic image segmentation,'' in
  \emph{Proceedings of the European Conference on Computer Vision}, 2018, pp.
  801--818.

\bibitem{dual}
J.~Fu, J.~Liu, H.~Tian, Y.~Li, Y.~Bao, Z.~Fang, and H.~Lu, ``Dual attention
  network for scene segmentation,'' in \emph{Proceedings of the IEEE Conference
  on Computer Vision and Pattern Recognition}, 2019, pp. 3146--3154.

\bibitem{ocr}
Y.~Yuan, X.~Chen, and J.~Wang, ``Object-contextual representations for semantic
  segmentation,'' \emph{arXiv preprint arXiv:1909.11065}, 2020.

\bibitem{self}
X.~Wang, R.~Girshick, A.~Gupta, and K.~He, ``Non-local neural networks,'' in
  \emph{Proceedings of the IEEE Conference on Computer Vision and Pattern
  Recognition}, 2018, pp. 7794--7803.

\bibitem{cem}
H.~Zhang, K.~Dana, J.~Shi, Z.~Zhang, X.~Wang, A.~Tyagi, and A.~Agrawal,
  ``Context encoding for semantic segmentation,'' in \emph{Proceedings of the
  IEEE Conference on Computer Vision and Pattern Recognition}, 2018, pp.
  7151--7160.

\bibitem{spn}
S.~Liu, S.~De~Mello, J.~Gu, G.~Zhong, M.-H. Yang, and J.~Kautz, ``Learning
  affinity via spatial propagation networks,'' in \emph{Proceedings of the
  Advances in Neural Information Processing Systems}, 2017, pp. 1520--1530.

\bibitem{cspn}
X.~Cheng, P.~Wang, and R.~Yang, ``Depth estimation via affinity learned with
  convolutional spatial propagation network,'' in \emph{Proceedings of the
  European Conference on Computer Vision}, 2018, pp. 103--119.

\bibitem{ade}
B.~Zhou, H.~Zhao, X.~Puig, S.~Fidler, A.~Barriuso, and A.~Torralba, ``Scene
  parsing through {ADE20K} dataset,'' in \emph{Proceedings of the IEEE
  Conference on Computer Vision and Pattern Recognition}, 2017, pp. 633--641.

\bibitem{dilation}
F.~Yu and V.~Koltun, ``Multi-scale context aggregation by dilated
  convolutions,'' \emph{arXiv preprint arXiv:1511.07122}, 2015.

\bibitem{acnet}
J.~Fu, J.~Liu, Y.~Wang, Y.~Li, Y.~Bao, J.~Tang, and H.~Lu, ``Adaptive context
  network for scene parsing,'' in \emph{Proceedings of the IEEE International
  Conference on Computer Vision}, 2019, pp. 6748--6757.

\bibitem{cpn}
C.~Yu, J.~Wang, C.~Gao, G.~Yu, C.~Shen, and N.~Sang, ``Context prior for scene
  segmentation,'' in \emph{Proceedings of the IEEE Conference on Computer
  Vision and Pattern Recognition}, 2020, pp. 12\,416--12\,425.

\bibitem{cfnet}
H.~Zhang, H.~Zhang, C.~Wang, and J.~Xie, ``Co-occurrent features in semantic
  segmentation,'' in \emph{Proceedings of the IEEE Conference on Computer
  Vision and Pattern Recognition}, 2019, pp. 548--557.

\bibitem{psanet}
H.~Zhao, Y.~Zhang, S.~Liu, J.~Shi, C.~Change~Loy, D.~Lin, and J.~Jia,
  ``{PSAN}et: Point-wise spatial attention network for scene parsing,'' in
  \emph{Proceedings of the European Conference on Computer Vision}, 2018, pp.
  267--283.

\bibitem{asymmetric}
Z.~Zhu, M.~Xu, S.~Bai, T.~Huang, and X.~Bai, ``Asymmetric non-local neural
  networks for semantic segmentation,'' in \emph{Proceedings of the IEEE
  International Conference on Computer Vision}, 2019, pp. 593--602.

\bibitem{apcnet}
J.~He, Z.~Deng, L.~Zhou, Y.~Wang, and Y.~Qiao, ``Adaptive pyramid context
  network for semantic segmentation,'' in \emph{Proceedings of the IEEE
  Conference on Computer Vision and Pattern Recognition}, 2019, pp. 7519--7528.

\bibitem{jointbilateral}
J.~Kopf, M.~F. Cohen, D.~Lischinski, and M.~Uyttendaele, ``Joint bilateral
  upsampling,'' \emph{ACM Transactions on Graphics}, vol.~26, no.~3, pp.
  96--es, 2007.

\bibitem{msg}
T.-W. Hui, C.~C. Loy, and X.~Tang, ``Depth map super-resolution by deep
  multi-scale guidance,'' in \emph{Proceedings of the European Conference on
  Computer Vision}, 2016, pp. 353--369.

\bibitem{hdsr}
C.~Guo, C.~Li, J.~Guo, R.~Cong, H.~Fu, and P.~Han, ``Hierarchical features
  driven residual learning for depth map super-resolution,'' \emph{IEEE
  Transactions on Image Processing}, vol.~28, no.~5, pp. 2545--2557, 2018.

\bibitem{sintel}
D.~J. Butler, J.~Wulff, G.~B. Stanley, and M.~J. Black, ``A naturalistic open
  source movie for optical flow evaluation,'' in \emph{Proceedings of the
  European Conference on Computer Vision}, 2012, pp. 611--625.

\bibitem{middle}
D.~Scharstein and R.~Szeliski, ``A taxonomy and evaluation of dense two-frame
  stereo correspondence algorithms,'' \emph{International Journal of Computer
  Vision}, vol.~47, no. 1-3, pp. 7--42, 2002.

\bibitem{zuo2019multi}
{Zuo et al.}, ``Multi-scale frequency reconstruction for guided depth map
  super-resolution via deep residual network,'' \emph{IEEE Transactions on
  Circuits and Systems for Video Technology}, vol.~30, no.~2, pp. 297--306,
  2019.

\bibitem{tgv}
J.~Park, H.~Kim, Y.-W. Tai, M.~S. Brown, and I.~S. Kweon, ``High-quality depth
  map upsampling and completion for {RGB-D} cameras,'' \emph{IEEE Transactions
  on Image Processing}, vol.~23, no.~12, pp. 5559--5572, 2014.

\end{thebibliography}

% biography section
% 
% If you have an EPS/PDF photo (graphicx package needed) extra braces are
% needed around the contents of the optional argument to biography to prevent
% the LaTeX parser from getting confused when it sees the complicated
% \includegraphics command within an optional argument. (You could create
% your own custom macro containing the \includegraphics command to make things
% simpler here.)
%\begin{IEEEbiography}[{\includegraphics[width=1in,height=1.25in,clip,keepaspectratio]{mshell}}]{Michael Shell}
% or if you just want to reserve a space for a photo:

\begin{IEEEbiography}[{\includegraphics[width=1in,height=1.25in,clip,keepaspectratio]{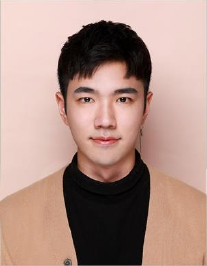}}]{Min-Cheol Sagong}
received his B.S. degree in Electrical Engineering from Korea University in 2018. He is currently pursuing his Ph.D. degree in Electrical Engineering at Korea University. His research interests are in the areas of digital signal processing, computer vision, and artificial intelligence.
\end{IEEEbiography}

\begin{IEEEbiography}[{\includegraphics[width=1in,height=1.25in,clip,keepaspectratio]{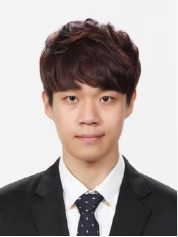}}]{Yoon-Jae Yeo}
received his B.S. degree in Electrical Engineering from Korea University in 2017. He is currently pursuing his Ph.D. degree in Electrical Engineering at Korea University. His research interests are in the areas of image processing, computer vision, and deep learning.
\end{IEEEbiography}

\begin{IEEEbiography}[{\includegraphics[width=1in,height=1.25in,clip,keepaspectratio]{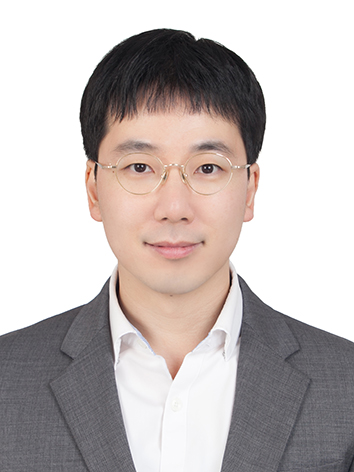}}]{Seung-Won Jung} (S’06-M’11-SM’19) received the B.S. and Ph.D. degrees in electrical engineering from Korea University, Seoul, Korea, in 2005 and 2011, respectively. He was a Research Professor with the Research Institute of Information and Communication Technology, Korea University, from 2011 to 2012. He was a Research Scientist with the Samsung Advanced Institute of Technology, Yongin-si, Korea, from 2012 to 2014. He was an Assistant Professor at the Department of Multimedia Engineering, Dongguk University, Seoul, Korea, from 2014 to 2020. In 2020, he joined the Department of Electrical Engineering at Korea University, where he is currently an Associate Professor. He has published over 70 peer-reviewed articles in international journals. He received the Hae-Dong young scholar award from the Institute of Electronics and Information Engineers in 2019. His current research interests include image processing and computer vision.
\end{IEEEbiography}

\begin{IEEEbiography}[{\includegraphics[width=1in,height=1.25in,clip,keepaspectratio]{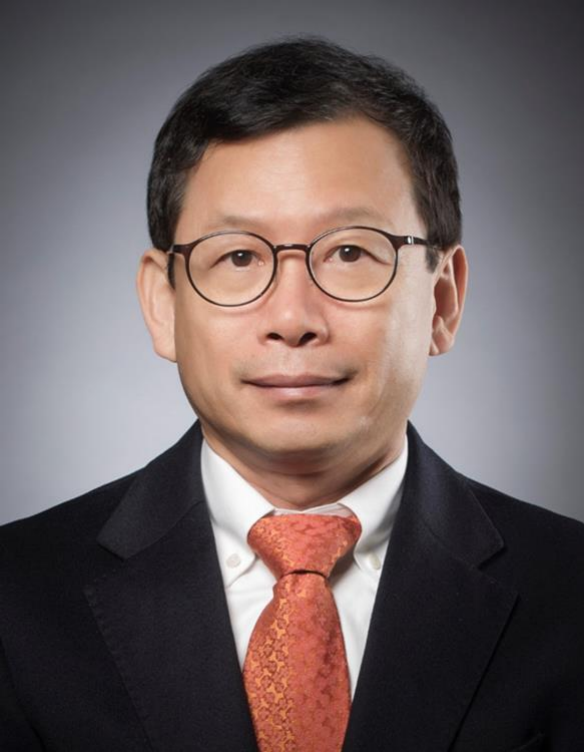}}]{Sung-Jea Ko}
(M’88-SM’97-F’12) received his Ph.D. degree in 1988 and his M.S. degree in 1986, both in Electrical and Computer Engineering, from State University of New York at Buffalo, and his B.S. degree in Electronic Engineering at Korea University in 1980. In 1992, he joined the Department of Electronic Engineering at Korea University where he is currently a Professor. From 1988 to 1992, he was an Assistant Professor in the Department of Electrical and Computer Engineering at the University of Michigan-Dearborn. He has published over 210 international journal articles. He also holds over 60 registered patents in fields such as video signal processing and computer vision. 

Prof. Ko received the best paper award from the IEEE Asia Pacific Conference on Circuits and Systems (1996), the LG Research Award (1999), and both the technical achievement award (2012) and the Chester Sall award from the IEEE Consumer Electronics Society (2017). He was the President of the IEIE in 2013 and the Vice-President of the IEEE CE Society from 2013 to 2016. He is a member of the National Academy of Engineering of Korea. He is a member of the editorial board of the IEEE Transactions on Consumer Electronics.
\end{IEEEbiography}

\end{document}